%% file: latex/acl_latex.tex
\documentclass[11pt]{article}

\usepackage[final]{acl}

\usepackage{times}
\usepackage{latexsym}

\usepackage[T1]{fontenc}

\usepackage[utf8]{inputenc}

\usepackage{microtype}

\usepackage{inconsolata}

\usepackage{graphicx}
\usepackage{arydshln}
\usepackage{array} 

\usepackage{tcolorbox}
\newtcolorbox{resq}{colback=yellow!4!white,colframe=green!75!black}

\usepackage{graphicx}

\usepackage{amsmath}
\usepackage{amssymb}
\usepackage{amsthm}

\usepackage{multirow}
\usepackage{booktabs}

\usepackage{pifont}

\usepackage{stfloats}
\usepackage{float}
\usepackage{caption}
\usepackage{soul}

\usepackage{pdflscape}

\usepackage{subfigure}
\usepackage{hyperref}
\usepackage{tabularx}
\usepackage{xspace}

\usepackage{color}

\usepackage{pgfplots}
\usepackage{times}
\usepackage{pgfplotstable}

\usepackage{colortbl}
\usepackage{xcolor}

\usepackage{arydshln}

\usepackage[ruled,vlined]{algorithm2e}

\usepackage{dashrule}

\title{SPS: Steering Probability Squeezing for Better Exploration in Reinforcement Learning for Large Language Models}

\author{
\textbf{Yifu Huo\textsuperscript{1},
Chenglong Wang\textsuperscript{1},
Ziming Zhu\textsuperscript{1},
Shunjie Xing\textsuperscript{1},
Peinan Feng\textsuperscript{1},
Tongran Liu\textsuperscript{2}
}, \\
\textbf{
Qiaozhi He\textsuperscript{1}
Tianhua Zhou\textsuperscript{3},
Xiaojia Chang\textsuperscript{3},
Jingbo Zhu\textsuperscript{1},
Zhengtao Yu\textsuperscript{4},
and Tong Xiao\textsuperscript{1}\thanks{Corresponding author.}}\\
\textsuperscript{1}Northeastern University, Shenyang, China \\
\textsuperscript{2}CAS Key Laboratory of Behavioral Science, Beijing, China \\
\textsuperscript{3}Independent Researcher, Beijing, China \\
\textsuperscript{4}Kunming University of Science and Technology, Kunming, China \\
\tt{ifnoct@gmail.com \quad xiaotong@mail.neu.edu.cn}
}

\begin{document}
\maketitle
\begin{abstract}
Reinforcement learning (RL) has emerged as a promising paradigm for training reasoning-oriented models by leveraging rule-based reward signals. However, RL training typically tends to improve single-sample success rates (\textit{i.e.}, Pass@1) while offering limited exploration of diverse reasoning trajectories, which is crucial for multi-sample performance (\textit{i.e.}, Pass@k). Our preliminary analysis reveals that this limitation stems from a fundamental \textit{squeezing effect}, whereby probability mass is excessively concentrated on a narrow subset of high-reward trajectories, restricting genuine exploration and constraining attainable performance under RL training. To address this issue, in this work, we propose \textbf{\underline{S}}teering \textbf{\underline{P}}robability \textbf{\underline{S}}queezing (SPS), a training paradigm that interleaves conventional RL with inverse reinforcement learning (IRL). SPS treats on-policy rollouts as demonstrations and employs IRL to explicitly reshape the induced trajectory distribution, thereby enhancing exploration without introducing external supervision. Experiments on five commonly used reasoning benchmarks demonstrate that SPS can enable better exploration and improve Pass@k. Beyond algorithmic contributions, we provide an analysis of RL learning dynamics and identify an empirical upper bound on Pass@k, shedding light on intrinsic exploration limits in RL-based reasoning models. Our findings suggest that alternating between RL and IRL offers an effective pathway toward extending the exploration capacity of reasoning-oriented large language models.
\end{abstract}

\section{Introduction}
In recent years, large language models (LLMs) have demonstrated impressive performance across a broad spectrum of foundational natural language processing (NLP) tasks, including text summarization, dialogue systems, and machine translation \cite{Stiennon2020LearningTS,wang-etal-2024-hybrid,Luo2025BeyondDL}.
Building on the advances, the research community has increasingly shifted its focus toward more challenging research frontiers, especially in reasoning and code generation \cite{Lightman2023LetsVS,Li2025STT}, and has even begun exploring the use of LLMs in the discovery of novel scientific theorems \cite{Georgiev2025MathematicalEA}.
As a result, exploration has emerged as a key capability of LLMs for future progress in these domains.

Motivated by the growing importance of exploration in reasoning-centric applications, contemporary LLM alignment methods have begun to explicitly incorporate exploration into the training pipeline. A simple and widely adopted strategy is to draw multiple samples per prompt to obtain a diverse set of candidate responses, where the model’s exploration capability is essential for ensuring output diversity \cite{Liu2023StatisticalRS,wang2024esrl}. However, such multi-sample strategies merely increase surface-level diversity by repeatedly sampling from an unchanged policy, without fundamentally enhancing the entropy of the underlying distribution, resulting in highly inefficient exploration \cite{Cui2025TheEM}.

This limitation has been further substantiated by recent empirical studies. For example, \citet{Yue2025DoesRL} demonstrate that although RL training substantially improves Pass@1 under large-scale sampling, the corresponding gains in Pass@k grow much more slowly, reflecting insufficient exploration of alternative reasoning trajectories. In essence, RL primarily improves sampling efficiency to boost single-sample success rates, rather than uncovering diverse trajectories that would meaningfully enhance multi-sample performance. To mitigate this sharpening effect and promote exploration, recent work has extended vanilla RL methods primarily along a common direction: explicitly counteracting entropy collapse to encourage broader exploration during RL training \cite{Liu2025ProRLPR,Cui2025TheEM}.

In this work, we advance this line of research by investigating a fundamental \textit{squeezing effect} in RL training \cite{Ren2024LearningDO}. This effect characterizes a systematic bias in probability mass redistribution. Specifically, negative gradients applied to low-probability responses fail to reallocate probability mass toward positively reinforced alternatives; instead, the removed mass is disproportionately absorbed by the greedy (\textit{i.e.}, already dominant) response. As a consequence, the output distribution becomes increasingly concentrated, exacerbating distributional sharpening rather than promoting exploration. Our preliminary analysis reveals that this squeezing effect constitutes an intrinsic limitation of exploration in RL-based training. Moreover, we provide a theoretical justification supporting this insight, formalizing how probability mass redistribution under standard RL objectives leads to progressive concentration (Please refer to Appendix~\ref{sec:proofs}).

Motivated by this analysis, we aim to explicitly enhance exploration by mitigating the squeezing effect. To this end, we propose \underline{\textbf{S}}teering \underline{\textbf{P}}robability \underline{\textbf{S}}queezing (SPS), an RL training approach that extends conventional RL by interleaving inverse reinforcement learning (IRL) stages. Our basic idea is that, following standard RL training, we employ an IRL to explicitly reshape the induced trajectory distribution, reallocating probability mass away from overly dominant responses toward under-explored but potentially valuable alternatives. Specifically, compared to vanilla RL, SPS periodically incorporates forward IRL updates \cite{sun2024inverse}, using only on-policy rollouts as demonstrations to avoid introducing external supervision or prior knowledge. Additionally, to further enhance exploration, we design an iterative SPS training strategy that repeatedly alternates between RL and IRL updates, enabling progressive redistribution of probability mass and preventing premature concentration of the policy. 

Our core contributions are threefold:
\begin{itemize}
    \item We conduct a preliminary analysis of the training dynamics in RL and identify an empirical upper bound on Pass@k. Our analysis results reveal the presence of a \textit{squeezing effect} in RL, which constrains exploration.
    \vspace{-1mm}
    \item Building on this analysis, we propose the SPS approach, which employs IRL to explicitly reshape the induced trajectory distribution, thereby facilitating enhanced exploration. Additionally, we introduce an iterative SPS training strategy to further enhance exploration.
    \item We evaluate SPS on five Olympiad-level mathematical benchmarks. The experimental results demonstrate consistent and substantial improvements in Pass@k, indicating that SPS effectively broadens exploration and facilitates the discovery of diverse reasoning trajectories. Notably, on the Qwen2.5-Math-1.5B model, SPS achieves a Pass@128 score of 63.33 on the \textsc{BrUMO} benchmark, representing an improvement of +10.00 points compared to the vanilla GRPO \cite{Shao2024DeepSeekMathPT}.
\end{itemize}

\section{Preliminaries}

\subsection{Task Formulation}
\paragraph{Enhancing LLM Reasoning via Constrained Data.} Given a finite set of reasoning questions $x$ with corresponding ground truth label ${l}$, the objective of enhancing LLM reasoning is to learn a policy that produces correct reasoning trajectories through on-policy rollouts.
During training, the policy iteratively samples multiple trajectories and receives corresponding outcome-level feedback extracted from the validator.
The validator can be written as
\begin{eqnarray}
    R(y,l) &=& \mathbb{I}[v(y)=l]
    \label{eq:outcome-level-feedback}
\end{eqnarray}
\noindent where $v(\cdot)$ denotes an extraction function that extracts the answer from response $y$.
In mathematical reasoning, the validator is commonly formulated as an indicator function, assigning a value of 1 when the extracted answer exactly matches the ground truth $l$, and 0 otherwise.

\paragraph{Exploration on Reasoning Tasks.}
In LLMs training, exploration refers to the ability of a learning process to expand the set of correct reasoning trajectories rather than simply reweighting partial existing patterns.
Formally, given a base policy $\pi_\mathrm{base}(\cdot)$ and a training policy $\pi_\theta(\cdot)$, exploration occurs if $\pi_\theta(\cdot)$ raises the probability to correct reasoning trajectories that are outside the high-likelihood region, thereby enlarging the boundary of the set of solvable problems.

\paragraph{Measurement of Exploration.}
Under our definition, effective exploration corresponds to expanding the set of problems that the model can successfully solve. 
To operationalize this notion, we adopt Pass@k as an estimation of the exploration. Pass@k is commonly defined as the expected maximum reward obtained from $k$ independently sampled responses for a given problem~\cite{Chen2025PasskTF}. Formally, it is computed as
\begin{eqnarray}
\kappa_k&=&\hspace{-0.3cm}\mathop{\mathbb{E}}_{\substack{_{(x,l)\sim D} \\ _{\{\hat{y_i}\}^k_{i=1}\sim \pi_\theta(\cdot|x)}}}\hspace{-0.3cm} [ \notag \\ &&\max(R(\hat{y_1},l),R(\hat{y_2},l),\cdots,R(\hat{y_k},l))] \notag \\
\end{eqnarray}
where $k$ is typically set to a relatively large value to reflect the model’s exploration capability. Following prior studies \cite{Ji2025LeanabellProverV2VR}, we set $k=128$ throughout our experiments.

\subsection{Group Relative Policy Optimization}
GRPO has emerged as one of the most widely adopted RL algorithms for training LLMs.
Compared to standard PPO \cite{Schulman2017ProximalPO}, GRPO estimates advantages using a group of $G$ rollouts rather than relying on a separate value network.
Despite this multi-sample formulation, the reward signal in the RLVR setting is binary (i.e., correct or incorrect), which allows the learning objective to be reformulated in a contrastive learning framework.
Building on this observation, \citeauthor{Wu2025ItTT} further decomposes the original objective into the following contrastive form:
\begin{eqnarray}
\mathcal{J}_{\text{GRPO}}(\theta) &=&  \sqrt{\text{Var}(x)} \Bigg( \mathop{\mathbb{E}}_{_{y^+\sim \pi_\theta^+(\cdot|x)}}\hspace{-0.3cm}\frac{\pi_\theta(y^+|x)}{|y^+|}  \notag \\
&&- \hspace{-0.3cm}\mathop{\mathbb{E}}_{_{y^-\sim \pi_\theta^-(\cdot|x)}}\hspace{-0.3cm}\frac{\pi_\theta(y^-|x)}{|y^-|} \Bigg)
\end{eqnarray}
where $\mathrm{Var}(\cdot)$ denotes the variance of the Bernoulli reward scores estimated from grouped samples, and $y^+$ and $y^-$ denote positively and negatively rewarded samples, respectively. $\pi_\theta^+(\cdot)$ and $\pi_\theta^-(\cdot)$ denote the positive and negative policy, respectively.

\begin{figure}[!t]
    \centering
    \subfigure[Influence of gradients on a balanced distribution.]{
        \includegraphics[width=0.45\textwidth]{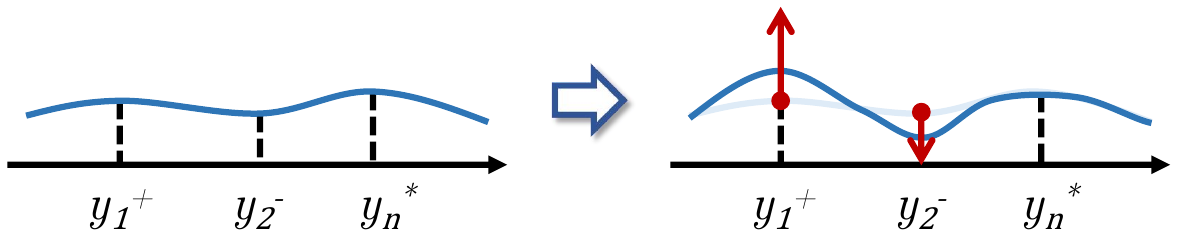}
        \label{fig:squeeze-a}
    }
    \vspace{-2mm}
    \subfigure[Influence of gradients on a peaky distribution.]{
        \includegraphics[width=0.45\textwidth]{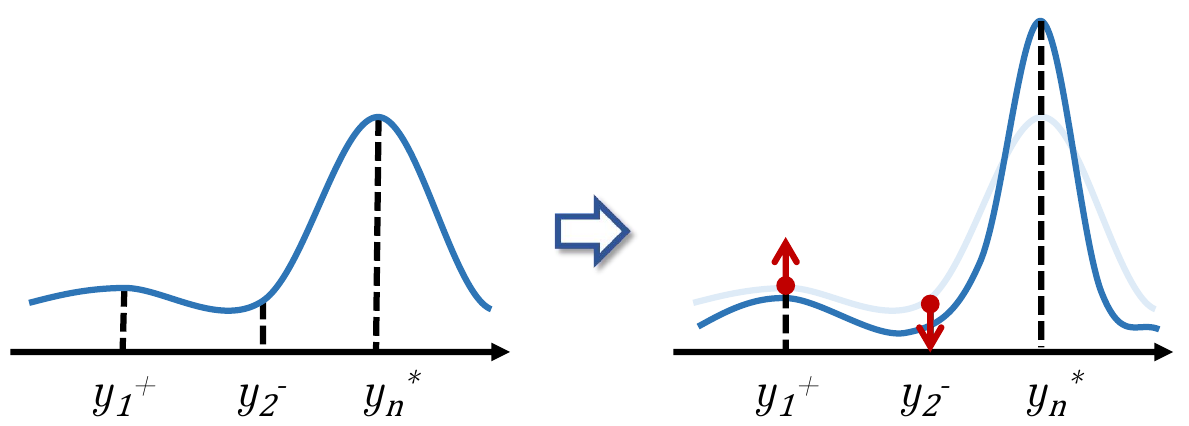}
        \label{fig:squeeze-b}
    }
    \caption{Illustration of \textit{squeezing effect}. $y_n^*$ denotes the sequence that dominates the output distribution (\textit{i.e.}, the sequence consistently sampled by greedy decoding). Subfigure (a) shows the normal RL case, where probability mass shifts along the gradient direction. Subfigure (b) shows that when the distribution is already imbalanced, the updates further concentrate probability mass into the dominant peak, a phenomenon referred to as the \textit{squeezing effect}. }
    \label{fig:squeezing-effect}
\end{figure}

\section{Preliminary Analysis}
Motivated by learning dynamics analyses~\cite{Ren2024LearningDO}, we hypothesize that the under-exploration issue in RL arises from an inherent squeezing effect induced by contrastive reward optimization. To validate this hypothesis, we conduct a two-stage analysis. First, we characterize how the squeezing effect emerges during RL training. Second, we explore how this effect restricts genuine exploration in reasoning tasks.

\subsection{Emergence of the Squeezing Effect in Reinforcement Learning}
\label{sec:squeezing-effect}

The \textit{squeezing effect} describes a phenomenon in which applying negative gradient updates to low-probability tokens paradoxically causes the model's output distribution to concentrate further on the most likely token. As illustrated in Figure~\ref{fig:squeeze-a}, when a policy model is trained with RL, its updates are jointly influenced by two opposing gradient components arising from the objective. Intuitively, the positive gradient increases the likelihood of positively rewarded samples, while the negative gradient suppresses the likelihood of negatively rewarded ones. However, this intuition breaks down under highly imbalanced output distributions, as shown in Figure~\ref{fig:squeeze-b}. When a small number of tokens already dominate the distribution, the probability mass removed from low-probability tokens is not redistributed evenly; instead, it is effectively \textit{squeezed} toward the dominant tokens, further amplifying their probabilities.

In fact, this counterintuitive behavior arises from the normalization property of the softmax function used in the model \cite{Ren2024LearningDO}. Specifically, when a negative update is applied to a token with negligible probability, the token itself is barely affected. Instead, the update primarily increases the softmax normalization constant, which reduces the normalized probabilities of nearly all tokens. For tokens that already dominate the distribution, however, this reduction is minimal in relative terms, causing their normalized probabilities to increase proportionally. As a result, probability mass progressively concentrates on the most likely token, leading to systematic sharpening of the output distribution and reduced diversity. A detailed theoretical proof of the squeezing effect is provided in Appendix~\ref{sec:proofs}.

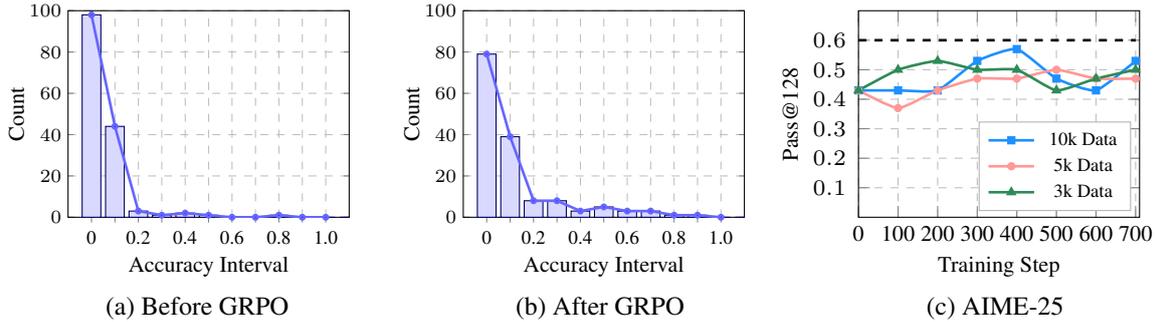
\begin{figure*}[!t]
    \centering
    \input{images/preliminary-study} 
    \vspace{-2mm}
    \caption{
       Partial results of the preliminary study. Subfigures (a) and (b) show the effect of GRPO on average question accuracy over the combined dataset. Subfigure (c) presents the dynamics of the Pass@128 metric during training, revealing an empirical boundary on exploration. More results can be found in Figure~\ref{fig:preliminary-study-appendix}.
    }
    \label{fig:preliminary-study}
\end{figure*}

\subsection{Impact of the Squeezing Effect on Exploration}
In this subsection, we analyze the impact of the squeezing effect on RL performance from an exploration perspective. Inspired by recent studies highlighting the importance of entropy and distributional sharpness in RL \cite{Cui2025TheEM,Yue2025DoesRL}, we argue that as the squeezing effect progressively reallocates probability mass toward already dominant tokens, the model’s output distribution becomes increasingly concentrated. A more closely related phenomenon is reported by \citeauthor{Tang2025RethinkingSP}, who observe that penalizing low-probability tokens suppresses unlikely outputs, thereby narrowing the distribution and reducing response diversity. This gradual loss of diversity directly constrains exploratory behavior during training, limiting the model’s ability to discover alternative and potentially superior reasoning trajectories.

Based on this insight, we conduct a preliminary study focusing on the evolution of solvable questions during RL training. Specifically, we fine-tune Qwen2.5-Math-7B on 10k questions sampled from \texttt{Openr1-Math-46k-8192} using GRPO, and evaluate intermediate checkpoints on a combined benchmark consisting of the five Olympiad-level datasets. 
For each question, we compute the average pass rate across multiple sampled responses and discretize these values into accuracy buckets, enabling us to examine how performance is distributed throughout the course of GRPO training. Figures~\ref{fig:preliminary-study}(a) and (b) present the histograms of the average Pass@1 accuracy distributions for the base model (denoted as \textit{Before GRPO}) and the best GRPO checkpoint (denoted as \textit{After GRPO}), respectively. As shown in the results, although GRPO introduces explicit exploration during training, the model does not consistently discover better trajectories for all questions. To further substantiate this observation, we also report Pass@128 results on AIME-25, where model checkpoints are evaluated every 100 training steps under different training data scales (3k, 5k, and 10k questions). Across all settings, we observe that increasing training steps does not lead to a monotonic improvement in Pass@128 performance, indicating that higher-quality trajectories are not continuously uncovered during training. Recent studies often attribute this phenomenon to entropy collapse in RL \cite{Cui2025TheEM}. However, rather than stopping at this surface-level explanation, we here probe a deeper underlying cause: \textit{the probability squeezing effect, which naturally acts as a key mechanism that can trigger entropy collapse}.

\section{Steering Probability Sequeezing}

\begin{figure*}
    \centering
    \includegraphics[width=0.99\linewidth]{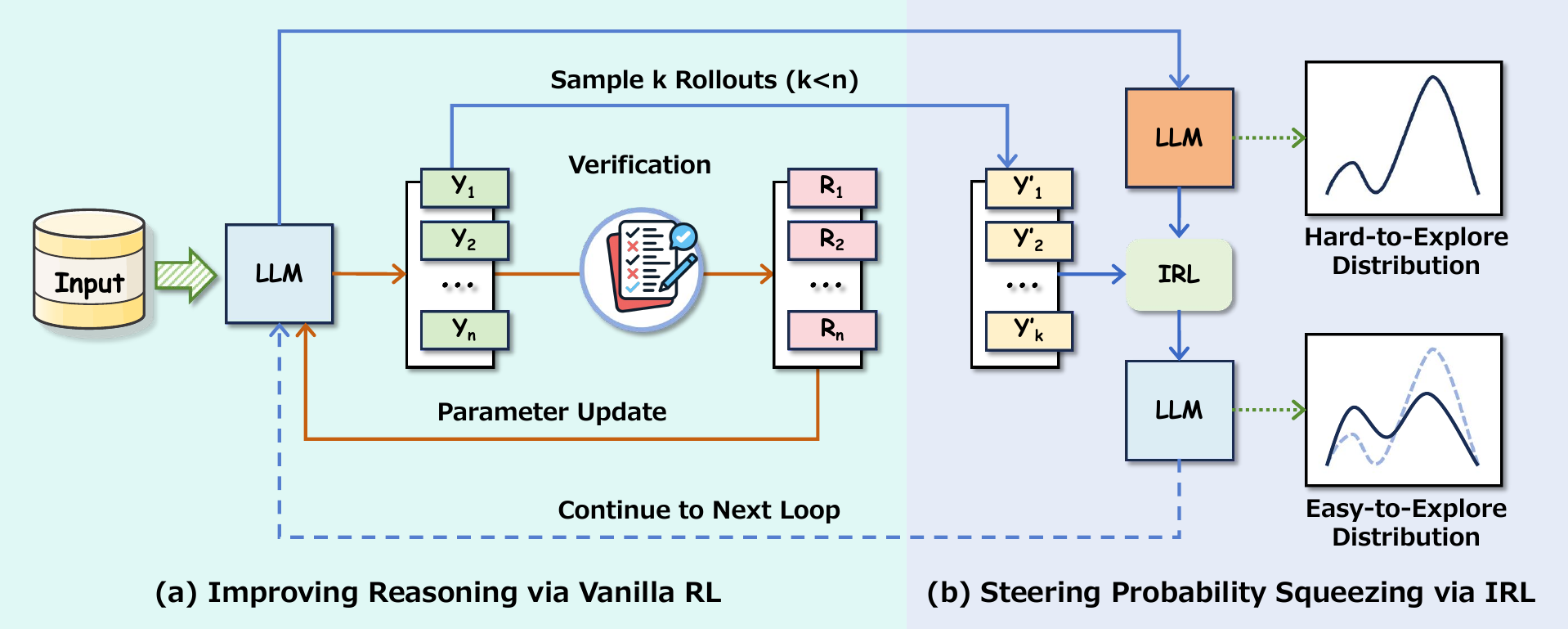}
    \vspace{-2mm}
    \caption{
    An overview of our SPS approach. Our training pipeline follows an iterative loop consisting of two complementary phases: (a) We first perform standard RL to explore the dataset and generate rollouts, from which a subset is sampled as demonstrations; (b) In the subsequent IRL phase, these demonstrations are leveraged to steer probability squeezing by reshaping the policy distribution. In practice, the two phases are interleaved to form a continual and unified training process.
    }
    \label{fig:main_figure}
\end{figure*}

From our preliminary analysis, we have established two key findings: 1) the squeezing effect occurs in RL, and 2) this squeezing effect limits the exploration.
These findings suggest that if we can steer this probability squeezing in a way that favors exploration, we could achieve improved RL performance.
To this end, we propose an SPS approach, which explicitly steers the probability squeezing phenomenon via interleaving on-policy RL with inverse RL. 
The basic idea of SPS is to ``redirect'' the misallocated probability mass during squeezing: instead of allowing it to converge to dominant greedy trajectories, we guide it toward under-explored regions that may contain better correct trajectories. 
The overview of SPS is shown in Figure~\ref{fig:main_figure}. We present the details of SPS in the following sections.

\subsection{Inverse Reinforcement Learning for Probability Redistribution}
Standard RL typically leads to probability squeezing, where excessive mass concentrates on a narrow set of high-reward trajectories. 
In this work, we adopt IRL as a principled mechanism to steer probability redistribution by matching desired occupancy patterns, rather than relying on ad hoc entropy regularization or reward reweighting.
In the IRL phase, we employ a forward-KL objective to reshape the policy’s output distribution. The loss function is defined as:
\begin{eqnarray}
    \mathcal{L}_\text{IRL}=-\mathbb{E}_{\substack{x\sim D \\ y' \sim Y'_x}}\text{KL}(\pi_\text{rollout}(y'|x)\Vert\pi(y'|x))    
\end{eqnarray}
where $x$ is sampled from the training dataset $D$, and $y'$ is sampled from a rollout set $Y_x'$, which is obtained by uniformly sampling from the responses generated during the vanilla RL phase. 
Here, $\pi_\text{rollout}$ denotes the empirical distribution over rollout completions, while $\pi$ represents the current policy. 
This objective encourages the policy to align with rollout-supported solution trajectories and thereby promotes broader exploration.

Crucially, during the IRL phases, we treat the rollouts generated by the current policy as the sole source of ``expert trajectories''.
Note that in this process, no external supervision, annotations, or domain knowledge is introduced.
As a result, SPS preserves the mass-covering nature of RL, while encouraging the model to explore beyond the narrow high-reward modes reinforced by standard RL.

\paragraph{Why Inverse Reinforcement Learning?}
IRL is no stranger to learn target distributions from demonstrations \cite{sun2024inverse,Sun2024SupervisedFA}. From a theoretical perspective, IRL enjoys an advantage that is particularly well-suited to our setting: it enables learning directly from example trajectories without explicitly specifying or constraining the policy through divergence-based regularization (e.g., KL-based approaches).
In our scenario, the goal is to explicitly control the probability squeezing phenomenon, specifically, to encourage probability mass to be redistributed toward under-explored yet potentially correct trajectories. Intuitively, this corresponds to steering the squeezing behavior in Figure~\ref{fig:preliminary-study}(b) to operate more like Figure~\ref{fig:preliminary-study}(a), where probability mass is concentrated around diverse high-quality trajectories rather than collapsing onto a few dominant ones. By incorporating IRL, we can directly leverage sampled rollouts as demonstrations to reshape the policy distribution, explicitly counteracting misallocated probability mass during squeezing. This allows the model to preserve exploration while still benefiting from reinforcement signals.

\paragraph{Low-Likelihood Trajectory Emphasis.}
Based on the analysis in Section~\ref{sec:squeezing-effect}, we observe that the squeezing effect primarily arises when optimization is dominated by negative samples with extremely low model likelihood. This observation suggests that explicitly increasing the influence of such low-likelihood solutions may help alleviate the squeezing phenomenon. Motivated by this insight, we propose Low-Likelihood Trajectory Emphasis (L2TE), a strategy that preferentially samples rollouts from trajectories with relatively low model likelihood. By amplifying the learning signal from these under-explored solutions, L2TE encourages broader exploration and counteracts excessive probability concentration. To ensure stable IRL training, we further augment each sampled batch with positive trajectories whenever the number of available negative samples is insufficient.

\subsection{Iterative Reinforcement Learning}
Since the IRL phase explicitly reshapes the model’s output distribution, it alleviates excessive distributional sharpening and thereby re-enables exploration within a fixed dataset. To further promote sustained exploration, we design a continually looped training strategy, as illustrated in Algorithm~\ref{alg:looped}. Specifically, we first fine-tune the base model using vanilla RL and collect the resulting rollouts. From these rollouts, we sample a small subset that balances exploration diversity and computational efficiency. The selected rollouts are then used to perform IRL on the reinforced policy, which reshapes the output distribution by redistributing probability mass away from overly dominant trajectories. This updated policy is subsequently fed back into the next RL phase. By iterating this RL–IRL loop, the model can continue to explore alternative solution trajectories even under constrained data conditions, progressively expanding the boundary of solvable problems rather than prematurely converging to a narrow set of greedy behaviors.

\begin{table*}[!t]
    \centering
    \resizebox{0.98\linewidth}{!}{
    \input{tables/main-results}}
    \vspace{-1mm}
    \caption{Performance comparison of RL methods across a set of reasoning benchmarks. Results are highlighted in bold when SPS outperforms vanilla GRPO, indicating enhanced exploration.}
    \label{tab:main-results}
\end{table*}

\section{Experiments}
\subsection{Experimental Setups}
\label{sec:ex-setup}

\paragraph{Dataset and Models.} 
Our experiments were conducted on Openr1-Math-46k-8192 \cite{Yan2025LearningTR}, which was a curated subset of OpenR1-Math-220k \cite{face2025open}. This subset removed excessively long or erroneous generations, ensuring that all questions were solvable. From this dataset, we constructed subsets of different scales (3k, 5k, and 10k) via uniform random sampling. For the base models, we conducted experiments using pretrained checkpoints from Qwen2.5-Math series, including 1.5B and 7B \cite{Yang2024Qwen25MathTR}.

\paragraph{Training Details.}
We implemented our method on top of \texttt{SWIFT}, using \texttt{vLLM} as the inference backend \cite{Kwon2023EfficientMM}. During the RL stages, we adopted a completion-level batch size of 128 and employed a reduced learning rate of 5e-7 to stabilize long-horizon exploration. Rollout generation was performed with a sampling temperature of 1.0, and we sampled 8 responses per prompt. Math-Verify\footnote{https://github.com/huggingface/Math-Verify} was used as the reward function without any additional format or length-based rewards. After the RL phase, we collected the generated rollouts and sampled three responses out of the eight completions for the IRL stages. To mitigate overfitting during IRL, we used a batch size of 512 and a learning rate of 5e-10. We performed four training steps per iteration to support extended exploration. All experiments were conducted on a cluster of $4\times8$ NVIDIA H100 GPUs.
More experimental details can be found in Appendix~\ref{app:implementsation_details}.

\paragraph{Evaluation.}
We implemented our SPS method on top of the GRPO algorithm, making GRPO \cite{Shao2024DeepSeekMathPT} our primary baseline. Additionally, we compared our method against several representative RL approaches, including DAPO \cite{Yu2025DAPOAO} and GSPO \cite{Zheng2025GroupSP}. Each baseline was implemented following the recommended configurations reported in the corresponding papers.  We evaluated our models across three challenging olympiad-level mathematical benchmarks to examine the boundary of solvable questions: \textsc{AIME} \cite{aime}, \textsc{BrUMO} \cite{brumo}, and HMMT \cite{hmmt}. For \textsc{AIME}, we considered both the 2024 and 2025 editions \citeyear{aime24,aime25}, and for \textsc{HMMT}, we evaluated both \textsc{HMMT-Feb} and \textsc{HMMT-Nov}. The evaluation was performed using \texttt{EvalScope}\footnote{https://github.com/modelscope/evalscope} \cite{evalscope_2024}, together with the benchmark data released by \citeauthor{balunovic_srimatharena_2025}. We reported Pass@128 and the average of Pass@1 (Avg@128) for all benchmarks, generating model outputs with a sampling temperature of 0.7.

\subsection{Main Results}

We report Pass@128 and Avg@128 for the best checkpoints within 700 training steps. 
The best checkpoint is selected according to Avg@128, as this metric reflects the convergence quality of RL training, as shown in Table~\ref{tab:main-results}. 
Our results demonstrate that SPS consistently outperforms all RL baselines on Pass@128, while maintaining comparable Avg@128 performance. 
This indicates that SPS improves both single-sample and multi-sample performance in a synchronized manner. 
Notably, SPS substantially increases Pass@128, implying that it effectively expands the exploration boundary. 
Remarkably, Qwen2.5-Math-1.5B achieves a Pass@128 score of 63.33 using only 3k training samples, highlighting the effectiveness of SPS in data-constrained settings.

The results also reveal an interesting pattern: \textit{the impact of GRPO varies with model scale}, and this trend is consistent across different data regimes. 
GRPO reduces Pass@128 for the 1.5B model, while improving it for the 7B model.
We hypothesize that this phenomenon is closely related to the base model’s initial output distribution. 
Smaller models (\textit{e.g.}, 1.5B) tend to overfit the training corpus, leading to a sharper distribution. 
GRPO aggravates this squeezing effect, thereby suppressing exploration. 
In contrast, larger models benefit from GRPO, which appears to enhance exploration by leveraging their richer internal knowledge.

\subsection{Ablation on Sampling Size}

\begin{figure}[!t]
    \centering
    \vspace{2mm}
    \input{images/ablation}
    \vspace{-2mm}
    \caption{
    Impact of the sampling size on SPS performance. The experiments are conducted on the Qwen2.5-Math-1.5B model.
    }
    \label{fig:ablation-sampling-size}
\end{figure}
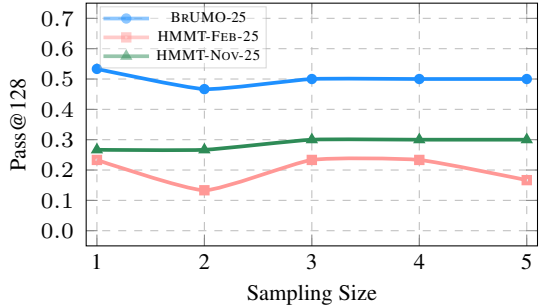

We further conducted an ablation study to investigate the effect of the sampling-size hyperparameter on training performance. Specifically, we applied SPS with varying sampling sizes on the 3k-sample dataset. Training was carried out for two epochs, and the results are summarized in Figure~\ref{fig:ablation-sampling-size}. The results demonstrate that model performance increased monotonically with larger sampling sizes. In practice, however, we balanced batch diversity against computational overhead and thus set the sampling size to three in all main experiments reported in this work.

\section{Related Works}

\paragraph{Reinforcement Learning for Large Reasoning Models.}
In the mainstream of current research, reasoning tasks are far from being low-hanging fruit. Unlike conventional NLP tasks, these \textit{logic-intensive} problems require multi-step inference and strict logical consistency, making them substantially more difficult to solve. Interestingly, despite their inherent complexity, the correctness of final answers can often be easily validated through rule-based procedures, such as exact matching or program execution \cite{Jiang2025CodeRLIC,Xie2025LogicRLUL,huo2025heal,wang2026msrl}. This property makes it feasible to train LLMs directly from outcome-level supervision, rather than relying on costly external annotations. Based on this observation, RL has emerged as an effective and explainable training paradigm for LLMs. Compared with RL from human feedback (RLHF), which relies on learned reward models to provide learning signals\cite{NEURIPS2022_b1efde53,zhou2024prior,wang2025gram,wang2025rovrm}, RLVR replaces human preference annotations with deterministic validators, enabling scalable and low-cost reward generation. However, recent studies have indicated that RLVR suffers from degraded exploration, as the learning process tends to concentrate probability mass on a narrow set of high-reward solutions, leading to a sharpened output distribution and limited discovery of novel reasoning patterns \citep{Yue2025DoesRL}.

\paragraph{Inverse Reinforcement Learning.}
IRL traditionally sought to infer an implicit reward function from expert demonstrations, framing learning as the recovery of objectives that rationalized observed behaviors \cite{sun2024inverse,sun2024improving,Deng2024FromNT}. In contrast to this classical setting, recent IRL-inspired approaches relaxed the reliance on external experts and instead operated on on-policy rollouts generated by the model itself \cite{wang2023learning}. From a self-supervising perspective, the model’s own trajectories serve as a proxy for demonstrations, allowing implicit reward functions to be extracted from its current behavior distribution \cite{zhang2021self}. Under this formulation, IRL no longer aims to exactly imitate an expert policy, but rather reshapes the reward or training signal to reweight model-generated trajectories, encouraging desirable solution patterns while preserving diversity. This perspective is particularly relevant in LLMs, where explicit rewards are often sparse or binary, and direct training tends to concentrate probability mass on a narrow set of high-reward outcomes. By leveraging on-policy rollouts as implicit supervision, IRL-style objectives provide a mechanism to smooth and redistribute the output distribution, complementing standard RL updates.

\section{Conclusion}
In this work, we have proposed SPS, an RL framework that interleaves on-policy RL with IRL to further enhance exploration. 
By learning from rollouts generated during the on-policy training phase, SPS can effectively mitigate the squeezing effect and significantly improve exploration compared with strong baselines across multiple olympiad-level reasoning benchmarks.  These results underscore the critical role of IRL, which is often overlooked as current research primarily emphasizes purely RL-based training. 
Additionally, this work highlights the importance of analyzing RL from the perspective of learning dynamics, providing a clearer explanation of the behavior and limitations of existing training paradigms.

\section*{Limitations}
While the proposed SPS approach provides a principled mechanism for steering probability mass to enhance exploration, several limitations warrant discussion. We discuss these limitations below:
\begin{itemize}
    \item Although our experiments demonstrate practical effectiveness in reasoning tasks, the empirical validation is restricted to a relatively small set of models, with Qwen2.5-Math serving as the primary benchmark due to its consistently strong performance.
    \item Although probability steering proves effective in mitigating the \textit{squeeze effect}, future work may explore more sophisticated mechanisms that more fully characterize and exploit the dynamics of reinforcement learning.
    \item Our current study does not analyze the inner states of policy models during training, leaving open questions regarding their interaction and relation to convergence behavior.
\end{itemize}


We acknowledge that we have not yet evaluated the method on larger-scale models. Due to computational constraints, our experiments focus on the 7B scale, which already allows us to study distributional concentration and exploration dynamics in a controlled setting.

\section*{Ethics Statement}

This work does not need ethical considerations.
The input of training is all from open-source data, and the output is also obtained based on open-source or commercial models.

\section*{Acknowledgments}
This work was supported in part by the National Natural Science Foundation of China (Nos. U24A20334 and 62276056), the Yunnan Fundamental Research Projects (No.202401BC070021), the Yunnan Science and Technology Major Project (No. 202502AD080014), the Fundamental Research Funds for the Central Universities (Nos. N25BSS054 and N25BSS094), and the Program of Introducing Talents of Discipline to Universities, Plan 111 (No.B16009). We would like to thank the anonymous reviewers and SPC for their valuable comments, which helped improve this paper.

\bibliography{latex/custom}

\appendix
\onecolumn

\begin{center}
    \Large \textbf{Supplementary Materials for SPS}
    \vspace{2mm}
\end{center}

\section{Proofs for Theoretical Results}
\label{sec:proofs}

\subsection{Derivation of the Squeezing Effect}
\textbf{Property 1.} \textit{The squeezing effect arises when negative gradient updates are applied to low-probability tokens, leading to a systematic sharpening of the model’s output distribution.}

\noindent \textbf{Proof: }This behavior is inherent to the normalization structure of the softmax function.
Let the model output distribution over the vocabulary be given by
\begin{eqnarray}
    p(i)=\frac{e^{z_i}}{Z}, 
\qquad 
Z=\sum_{j} e^{z_j}
\end{eqnarray}
where $z_i$ denotes the logit associated with token $i$.
Consider a token $m$ that receives a negative logit update during training,
\begin{eqnarray}
    z_m \leftarrow z_m + \eta,
\qquad 
\eta<0
\end{eqnarray}
\noindent which yields the updated distribution
\begin{eqnarray}
p'(i)=\frac{e^{z_i}}{Z'},
\qquad
Z' = e^{z_m+\eta}+\sum_{j\neq m} e^{z_j}
\end{eqnarray}
For any token $j\neq m$, we may express the updated probability in terms of the original distribution as
\begin{eqnarray}
    p'(j)
=
\frac{p(j)}{1+p(m)\left(e^{\eta}-1\right)}
\end{eqnarray}
The squeezing effect typically arises when the distribution satisfies $p(m)\ll1$. 
Performing a first-order expansion then gives
\begin{eqnarray}
    p'(j)
\approx
p(j)\left[1-p(m)\left(e^{\eta}-1\right)\right]
\end{eqnarray}
Since $\eta < 0$ implies $e^\eta-1<0$, it follows that
\begin{eqnarray}
    p'(j)<p(j), 
\qquad 
\forall j\neq m
\end{eqnarray}
That is, the normalized probabilities of almost all tokens decrease simultaneously. 
However, the best-probability token, being the dominant term in the distribution, which experiences the smallest relative decrease, implying
\begin{eqnarray}
    \max_i p'(i) > \max_i p(i)
\end{eqnarray}
Thus, the probability mass is progressively concentrated toward the most likely token, and the output distribution becomes increasingly peaked. 
This phenomenon is referred to as the squeezing effect \cite{Ren2024LearningDO}.

\subsection{Squeezing Effect at the Sequence Level}
The previous analysis establishes that penalizing low-probability tokens induces probability mass to concentrate toward the modal token due to the normalization structure of the softmax function. 
We now generalize this reasoning to sequence-level probability distributions, which are central to policy optimization in language model training.

\noindent \textbf{Property 2.} \textit{The squeezing effect arises when negative gradient updates are applied to low-probability sequences, leading to a systematic sharpening of the model’s output distribution.}

\noindent \textbf{Proof: }Let a sequence be denoted by
\begin{eqnarray}
    y=\{y_1,...,y_T\}
\end{eqnarray}
\noindent and let the model define the joint probaiblity
\begin{eqnarray}
    p(y)=\prod_{t=1}^T p(y_t \mid y_{<t})
\end{eqnarray}
\noindent where each conditional distribution is parameterized by a softmax over logits $z_t$.
Suppose that a particular sequence $y^-$ receives a negative gradient update under the training objective, effectively reducing its log-probability. 
This corresponds to a logit-space update of the form
\begin{eqnarray}
    \log p(y^{-}) \leftarrow \log p(y^{-}) + \eta,
\qquad \eta < 0
\end{eqnarray}
At the sequence level, the normalized model distribution over all candidate sequences $\mathcal{Y}$ may be represented as
\begin{eqnarray}
    P(y)=\frac{\exp(\log p(y))}{\sum_{y'\in\mathcal{Y}} \exp(\log p(y'))}
\end{eqnarray}
After the update to $y^-$, the new distribution becomes
\begin{eqnarray}
    P'(y)
=
\frac{\exp(\log p(y))}{
\exp(\log p(y^{-})+\eta)
+
\sum_{y'\neq y^{-}}
\exp(\log p(y'))
}
\end{eqnarray}
For any $y \neq y^-$, we obtain
\begin{eqnarray}
    P'(y)
=
\frac{P(y)}{
1 + 
P(y^{-})
\left(e^{\eta}-1\right)
}
\end{eqnarray}

\noindent If the penalized sequence is already extremely unlikely, \textit{i.e}.

\begin{eqnarray}
    P(y^{-}) \ll 1
\end{eqnarray}

\noindent then a first-order expansion yields

\begin{eqnarray}
    P'(y)
\approx
P(y)\left[
1 - 
P(y^{-})
\left(e^{\eta}-1\right)
\right]
\end{eqnarray}

\noindent Since $\eta < 0$ implies $e^\eta-1<0$, it follows that
\begin{eqnarray}
    P'(y)<P(y),
\qquad \forall y\neq y^{-}.
\end{eqnarray}

\noindent Thus, the normalized probability of nearly every sequence decreases simultaneously.

\noindent Let
\begin{eqnarray}
    y^\star = \arg\max_y P(y)
\end{eqnarray}

\noindent $y^\star$ denotes the most probable sequence.
Because this sequence dominates the distribution, its relative decrease under normalization is smallest. 
Consequently,
\begin{eqnarray}
    \max_y P'(y)
>
\max_y P(y),
\end{eqnarray}

\noindent implying that probability mass becomes increasingly concentrated on $y^\star$.

\section{RLVR Algorithms}
In this section, we enumerate the RLVR algorithms referred in this paper.
\subsection{Group Relative Policy Optimization (GRPO)}
In RLVR, GRPO has become one of the most widely used RL algorithms for LLM training.
GRPO maximizes expected rewards by increasing the likelihood of higher-reward samples within a group, while normalizing each sample’s advantage by the group’s average reward and variance. 
It removes the critic network and instead computes a relative advantage inside each sampled group, then applies a PPO-style clipped objective to stabilize updates.
The loss function of GRPO can be written as
\begin{align}
\mathcal{J}_\text{GRPO}(\theta)
=
\frac{1}{G}
\sum_{i=1}^{G}
\frac{1}{|y_i|}
\sum_{t=1}^{|y_i|}
\min\left(
w_{i}(\theta)\,\widehat{A}_{i},\;
\mathrm{clip}\!\left(w_{i}(\theta),\,1-\varepsilon,\,1+\varepsilon\right)\widehat{A}_{i}
\right)
\end{align}

\noindent where $w_i(\theta)$ denotes an importance ratio, which can be computed as
\begin{equation}
    w_{i}(\theta)=\frac{\pi_{\theta}(y_{i} \mid x)}{\pi_{\theta_{\text{old}}}(y_{i} \mid x)}
\end{equation}
Specially, GRPO computes the advantages $\widehat{A}_i$ by normalizing rewards within a group of responses.
In RLVR, we use outcome-level feedback given in Equation~\ref{eq:outcome-level-feedback} as reward, therefore the advantages are computed as:
\begin{eqnarray}
   \widehat{A}_i &=& \frac{R_i - \text{mean}(\{R_1, \dots, R_G\})}{\text{std}(\{R_1, \dots, R_G\})}
   \label{eq:group-normalize}
\end{eqnarray}

\subsection{Dynamic Sampling Policy Optimization (DAPO)}
To stabilize RL training, \citet{Yu2025DAPOAO} propose DAPO.
In DAPO, the clipping range is asymmetric: the lower bound remains restrictive to control instability, while the upper bound is relaxed to encourage exploration of low-probability tokens. 
Unlike GRPO, gradients are computed at the token level and averaged across all tokens in all sampled responses. 
Prompts for which all sampled responses are correct or all are incorrect are filtered out so that every retained prompt contributes a non-zero learning signal.
The loss function of DAPO can be written as

\begin{equation}
\begin{aligned}
\mathcal{J}_\text{DAPO}(\theta)
=
\frac{1}{\sum_{i=1}^{G}|y_i|}
\sum_{i=1}^{G}
\sum_{t=1}^{|y_i|}
\min\Big(
r_{i,t}(\theta)\,\widehat{A}_{i},\;
\mathrm{clip}\!\big(
r_{i,t}(\theta),\,
1-\varepsilon_{\text{low}},\,
1+\varepsilon_{\text{high}}
\big)\widehat{A}_{i}
\Big),
\\[2mm]
\text{s.t.}\quad
0<
\big|\{y_i \mid R(y_i,l) = 1\}\big|
<
G
\end{aligned}
\end{equation}

\noindent where $r_{i,t}(\theta)$ denotes a token-level importance ratio, as follows:

\begin{equation}
    r_{i,t}(\theta)=\frac{\pi_{\theta}(y_{i,t} \mid x, y_{i<t})}{\pi_{\theta_{\text{old}}}(y_{i,t} \mid x,y_{i<t})}
\label{eq:advantage_calculation}
\end{equation}

\subsection{Group Sequence Policy Optimization (GSPO)}
GSPO optimizes a sequence-level clipped objective, where each response’s normalized reward (advantage) is weighted by its sequence-likelihood ratio between the current and old policy. 
In essence, it performs PPO-style clipping at the whole-sequence level, aligning off-policy correction and optimization with the sequence-level reward.
The loss function of GSPO can be written as
\begin{align}
\mathcal{J}_\text{GSPO} (\theta)
=
\frac{1}{G} \sum_{i=1}^{G}
\min \left(
s_{i}(\theta)\,\widehat{A}_{i},\;
\mathrm{clip}\!\left( w_{i}(\theta),\, 1-\varepsilon,\, 1+\varepsilon \right)\widehat{A}_{i}
\right)
\end{align}

\noindent while its importance ratio $s_i(\theta)$ is differently computed as
\begin{eqnarray}
    s_{i}(\theta) = \left( \frac{ \pi_{\theta} (y_i | x) }{ \pi_{\theta_\text{old}} (y_i | x)} \right)^{\frac{1}{|y_i|}}
=
\exp \left( \frac{1}{|y_i|} \sum_{t=1}^{|y_i|} \log \frac{ \pi_{\theta} (y_{i,t} | x, y_{i,<t}) }{ \pi_{\theta_\text{old}} (y_{i,t} | x,y_{i,<t})} \right)
\end{eqnarray}

\noindent Therefore, GSPO applies clipping to entire responses instead of individual tokens to exclude the overly ``off-policy'' samples from gradient estimation, which matches both the sequence-level rewarding and optimization \cite{Zheng2025GroupSP}.

\section{Implementation Details}
\label{app:implementsation_details}
\subsection{Iterative SPS}
\label{sec:iter-sps-implement}
We propose the pseudo code of iterative SPS in Algorithm~\ref{alg:looped}.
The Iterative SPS algorithm iteratively enhances a base policy by alternating exploration and distribution reshaping: starting from the initial policy, it updates the policy on the dataset using vanilla RL to encourage exploration, collects grouped rollouts, samples a subset emphasizing low-likelihood or under-explored trajectories, and then applies IRL on this subset to reshape the policy distribution and mitigate probability squeezing, producing a more balanced and robust enhanced policy for exploration.

\input{algorithm/looped}

In implementation, the rollout distribution $\pi_\text{rollout}$ is instantiated through a degenerate discrete distribution over the sampled responses.
Under this construction, the forward-KL objective can be reformulated in a manner that closely resembles a cross-entropy style penalty, as previously discussed in related literature \citep{Sun2024SupervisedFA}. 
Interestingly, despite the apparent simplicity of this surrogate, empirical evidence suggests that it nevertheless facilitates an effective enlargement of the exploration region, even in the absence of explicit external guidance.

Furthermore, while any single low-probability response is unlikely to be sampled, the total number of such responses is exceedingly large. Consequently, the probability of obtaining at least one low-likelihood trajectory within a batch remains high. To better emphasize these low-probability trajectories, we sample from the lower quantile of trajectories in each batch.

\subsection{Hyperparameter Setting}
We conduct our main experiments using several RL algorithms, including GRPO \cite{Shao2024DeepSeekMathPT}, DAPO \cite{Yu2025DAPOAO}, and GSPO \cite{Zheng2025GroupSP}. 
The algorithm-specific hyperparameters are summarized in Table~\ref{tab:parameter-setting}. 
For GRPO, we adopt a fixed group size across all baseline methods in order to balance the exploration capacity and computational cost. 
For DAPO and GSPO, we employ the recommended default configurations provided in \texttt{SWIFT}, which have been previously validated in practical deployments.

\begin{table}[!t]
    \centering
    \resizebox{0.45\linewidth}{!}{

\input{tables/hyperparameters}}
    \vspace{-1mm}
    \caption{Hyperparameter setting of applied RL methods}
    \vspace{-3mm}
    \label{tab:parameter-setting}
\end{table}


\begin{figure*}[!t]
    \centering
    \input{images/appendix_acc_bucket} 
    \vspace{-2mm}
    \caption{
       Accuracy distribution variation of Qwen2.5-Math-1.5B
    }
    \vspace{-2mm}
    \label{fig:preliminary-study-appendix}
\end{figure*}

\section{Results on Other Backbone Models and Benchmarks}
To further validate the generalizability of our method, we extend the main experiment to incorporate DeepSeek-R1 \cite{DeepSeekAI2025DeepSeekR1IR} as our additional backbone model. 
Furthermore, we extend our evaluation to other non-mathematical benchmarks, such as MMLU \cite{Hendrycks2020MeasuringMM} and GPQA-Diamond \cite{Rein2023GPQAAG}, the extended results are listed as Table \ref{tab:additional-result}

\begin{table}[!t]
    \centering
    \resizebox{1.0\linewidth}{!}{
    \input{tables/additional-res}}
    \vspace{-1mm}
    \caption{Additional experimental results on DeepSeek-R1. The best results for each group are in \textbf{bold}.  The second-best results for each group are with \underline{underline}.}
    \vspace{-3mm}
    \label{tab:additional-result}
\end{table}

The extended results consistently show that SPS maintains its effectiveness across different model families and non-mathematical domains, supporting the general applicability of our method beyond competition-level mathematics.

\section{Further Analysis}
\subsection{Analysis of Diversity}
As Pass@K (\textit{e.g.}, Pass@128) is only an indirect proxy for exploration. Improvements in Pass@K may stem from multiple factors, including enhanced trajectory diversity. Nevertheless, prior work \cite{Yue2025DoesRL} suggests an intrinsic relationship between exploration dynamics and Pass@K metrics, as broader policy support generally increases the probability of sampling correct reasoning paths within a finite budget. In this sense, while Pass@K is not a direct diversity measure, it remains behaviorally correlated with exploration capacity.

At the same time, we fully agree that explicit diversity metrics provide more direct evidence. Following this suggestion, we compute trajectory-level similarity (lower indicates higher diversity), with results shown in Table \ref{tab:div-analysis}.

\begin{table}[!t]
    \centering
    \resizebox{0.4\linewidth}{!}{
    \input{tables/div-analysis}}
    \vspace{-1mm}
    \caption{Diversity comparison across different methods.}
    \vspace{-3mm}
    \label{tab:div-analysis}
\end{table}

The results show that SPS yields lower trajectory similarity, indicating increased reasoning diversity compared to both the base model and GRPO. This empirical evidence complements the Pass@K improvements and provides more direct support for our exploration claim.

\subsection{A Cost-Benefit Analysis on SPS}
As a multi-stage training method, SPS inevitably adds computational overhead and engineering complexity compared to vanilla RLVR methods. 
However, in practice, the IRL stage contributes only a negligible fraction of the total training time and does not introduce any substantial computational overhead. 
Empirically, the overall runtime is dominated by the RL rollout and policy optimization stage, whereas the IRL update introduces only marginal computational overhead. 
To quantify this, we measure the training time by training a 1.5B-parameter LLM on 3k prompts, isolating the respective stages. The measured training time is summarized in Table \ref{tab:time-cost}.

\begin{table}[!t]
    \centering
    \resizebox{0.3\linewidth}{!}{
    \input{tables/cost-analysis}}
    \vspace{-1mm}
    \caption{Time cost comparison across different stages.}
    \vspace{-3mm}
    \label{tab:time-cost}
\end{table}

As shown, the IRL stage accounts for only about 3\% of the total time per iteration, which is minor compared to the RL stage. Therefore, although the pipeline is conceptually multi-stage, the additional computational cost introduced by IRL is marginal and does not constitute a practical bottleneck.

\subsection{Diagnostics for \textit{Squeezing Effect}}
We examine probability dynamics by collecting responses generated via greedy decoding and computing the average log-probability of the generated trajectories under the current policy. Intuitively, excessive probability squeezing manifests as over-concentration of mass on a narrow subset of trajectories, typically reflected in inflated log-probability magnitudes relative to the base model. We evaluate several algorithms on AIME 2024 and 2025, with results shown in Table \ref{tab:squeeze-analysis}.

\begin{table}[!t]
    \centering
    \resizebox{0.4\linewidth}{!}{
    \input{tables/squeeze-analysis}}
    \vspace{-1mm}
    \caption{The average log-probability of the generated trajectories under different optimization methods.}
    \vspace{-3mm}
    \label{tab:squeeze-analysis}
\end{table}

Compared to GRPO and GSPO, SPS maintains log-probability levels much closer to the base model, indicating that it avoids aggressively concentrating probability mass on a small subset of trajectories. In contrast, GRPO and GSPO exhibit noticeably higher (less negative) log-probabilities, suggesting stronger probability squeezing.

These results provide direct empirical evidence that SPS mitigates probability squeezing while preserving performance gains. We will incorporate this diagnostic analysis into the revised manuscript to clarify the mechanism underlying SPS.

\section{Discussion}

\subsection{How SPS influence the reasoning?}
SPS does not merely flatten the output distribution at the logit level. If its effect were equivalent to temperature scaling or entropy regularization, we would expect uniform entropy increases without meaningful changes in internal representations. However, SPS operates on trajectory-level objectives and reweights complete reasoning paths, which propagates gradients through intermediate transformer layers rather than only adjusting the final projection head. Empirically, the gains in high-K metrics exceed what would be predicted from Pass@1 improvements under an independent sampling assumption, suggesting reduced inter-sample redundancy rather than simple probability smoothing. Conceptually, post-hoc logit flattening cannot induce new reasoning modes, whereas SPS reshapes probability mass across semantically distinct trajectories.

\subsection{Why the \textit{Degenerate Discrete Distribution} works well?}
The \textit{degenerate discrete distribution} is simply the empirical distribution over RL rollouts, \textit{i.e.}, a Monte Carlo estimator of the improved policy. 
Since the IRL stage only needs to match the relative structure within the sampled support, rather than reconstructing a continuous density as this empirical approximation is sufficient in practice. The target distribution is rollout-induced, so the empirical measure is a consistent surrogate. While a very small batch may increase variance, performance does not collapse in practice. This is partly due to the \textit{rare-but-many} effect, although individual low-probability trajectories are hard to sample, their combinatorial cardinality is large, so typical batches still contain diverse underrepresented modes. Moreover, the IRL update is conservative (small learning rate), which prevents overfitting to sampling noise.

\end{document}

%% file: images/preliminary-study.tex
\definecolor{blue1}{RGB}{29,143,255}
\definecolor{blue2}{RGB}{74,165,255}
\definecolor{blue3}{RGB}{142,199,255}
\definecolor{mygreen}{RGB}{46,139,87}
\definecolor{myred}{RGB}{255,152,150}
\definecolor{myblue}{RGB}{30,144,255}

\begin{tikzpicture}
       \scriptsize{    

\begin{axis}[
    ybar, 
    bar width=7pt,
    ymajorgrids,
    xmajorgrids,
    grid style=dashed,
    at={(0,0)},
    height=.27\textwidth,
    width=.33\textwidth,
    xmin=-0.1,
    xmax=1.1,
    ymin=0,
    ymax=100,
    xtick={0, 0.1, 0.2, 0.3, 0.4, 0.5, 0.6, 0.7, 0.8, 0.9, 1.0},
    xticklabels={0, , 0.2, , 0.4, , 0.6, , 0.8, , 1.0, }, 
    xticklabel style={/pgf/number format/fixed, font=\tiny, scale=1.2},
    xtick pos=bottom,
    major tick length=2pt,
    ytick={0, 20, 40, 60, 80, 100},
    scaled x ticks=false,
    ylabel=\scriptsize{Count},
    ylabel style={yshift=-5.0ex,scale=1.2},
    xlabel=\scriptsize{Accuracy Interval},
    xlabel style={yshift=1ex,scale=1.2},
    legend style={column sep=.45em, draw=black!30, legend columns=1},
    legend pos=north east
    ]
    
    \addplot+[ybar, fill=blue!15, draw=blue!40!black] coordinates {
        (0.0, 98)
        (0.1, 44)
        (0.2, 3)
        (0.3, 1)
        (0.4, 2)
        (0.5, 1)
        (0.6, 0)
        (0.7, 0)
        (0.8, 1)
        (0.9, 0)
        (1.0, 0)
        (1.1, 0)
    };
    \addplot+[
        sharp plot,       
        draw=blue!60,      
        fill=none,
        line width=1.0pt, 
        mark=*,           
        mark options={fill=blue!80, draw=none, scale=0.4} 
    ] coordinates {
        (0.0, 98)
        (0.1, 44)
        (0.2, 3)
        (0.3, 1)
        (0.4, 2)
        (0.5, 1)
        (0.6, 0)
        (0.7, 0)
        (0.8, 1)
        (0.9, 0)
        (1.0, 0)
    };
\end{axis}

\node [anchor=center] at (1.7cm,-1.2cm) {\scalebox{1.4}{(a) Before GRPO}};

\begin{axis}[
    ybar,
    bar width=7pt,
    ymajorgrids,
    xmajorgrids,
    grid style=dashed,
    at={(5.2cm,0)}, 
    height=.27\textwidth,
    width=.33\textwidth,
    xmin=-0.1,
    xmax=1.1,
    ymin=0,
    ymax=100,
    xtick={0, 0.1, 0.2, 0.3, 0.4, 0.5, 0.6, 0.7, 0.8, 0.9, 1.0},
    xticklabels={0, , 0.2, , 0.4, , 0.6, , 0.8, , 1.0, },
    xticklabel style={/pgf/number format/fixed, font=\tiny, scale=1.2},
    xtick pos=bottom,       
    major tick length=2pt,  
    ytick={0, 20, 40, 60, 80, 100},
    scaled x ticks=false,
    ylabel=\scriptsize{Count},
    ylabel style={yshift=-5.0ex,scale=1.2},
    xlabel=\scriptsize{Accuracy Interval},
    xlabel style={yshift=1ex,scale=1.2},
    legend style={column sep=.45em, draw=black!30, legend columns=1},
    legend pos=north east
    ]
    
    \addplot+[ybar, fill=blue!15, draw=blue!40!black] coordinates {
        (0.0, 79)
        (0.1, 39)
        (0.2, 8)
        (0.3, 8)
        (0.4, 3)
        (0.5, 5)
        (0.6, 3)
        (0.7, 3)
        (0.8, 1)
        (0.9, 1)
        (1.0, 0)
        (1.1, 0)
    };
    \addplot+[
    sharp plot,       
    draw=blue!60,      
    fill=none,
    line width=1.0pt, 
    mark=*,           
    mark options={fill=blue!80, draw=none, scale=0.4} 
] coordinates {
        (0.0, 79)
        (0.1, 39)
        (0.2, 8)
        (0.3, 8)
        (0.4, 3)
        (0.5, 5)
        (0.6, 3)
        (0.7, 3)
        (0.8, 1)
        (0.9, 1)
        (1.0, 0)
};

\end{axis}
}
    \node [anchor=center] at (7.0cm,-1.2cm) {\scalebox{1.4}{(b) After GRPO}};
    \scriptsize{    
    \begin{axis}
        [
        ymajorgrids,
        xmajorgrids,
        grid style=dashed,
        at={(10.4cm,0)},
        height=.27\textwidth,
        width=.33\textwidth,
        xmin=0,
        xmax=7.1,
        xtick={0,1,...,7},
        xticklabels={0,100,...,700}, 
        xticklabel style={/pgf/number format/fixed}, 
        scaled x ticks=false,
        ymin=0,
        ymax=0.7,
        ytick={0.1,0.2,...,0.6},
        xtick=data,
        x tick label style={/pgf/number format/fixed,
            /pgf/number format/fixed zerofill,
        /pgf/number format/precision=1, scale=1.2},
        y tick label style={/pgf/number format/fixed, xshift=-0.5ex,
            /pgf/number format/fixed zerofill,
        /pgf/number format/precision=1, scale=1.2},
        ylabel=\scriptsize{Pass@128},
        ylabel style={yshift=-3.0ex,scale=1.2},
        xlabel=\scriptsize{Training Step},
        xlabel style={yshift=1ex,scale=1.2},
        legend style = {
                        column sep=.45em,
                        draw=black!30,
                        legend columns=1},
        legend pos=south east
                        ]

    \addplot+[
    myblue,
    mark=none,
    mark size=0.8pt,
    line width=1.0pt,
    smooth, tension=0.3,
    mark=square,
    mark options={fill=myblue,draw=myblue,line width=1.5pt},
]
    table [
        x=X, y=Y
    ] {
        X   Y   
        0       0.43   
        1       0.43 
        2       0.43 
        3       0.53  
        4       0.57  
        5       0.47  
        6       0.43  
        7       0.53 
    };
    \addlegendentry[scale=1]{{10k Data}};

    \addplot+[
    myred,
    mark=none,
    mark size=0.8pt,
    line width=1.0pt,
    smooth, tension=0.3,
    mark=*,
    mark options={fill=myred,draw=myred,line width=1.5pt},
]
    table [
        x=X, y=Y
    ] {
        X   Y   
        0       0.43   
        1       0.37 
        2       0.43 
        3       0.47  
        4       0.47  
        5       0.50  
        6       0.47  
        7       0.47 
    };
    \addlegendentry[scale=1]{{5k Data}};

     \addplot+[
    mygreen,
    mark=none,
    mark size=0.8pt,
    line width=1.0pt,
    smooth, tension=0.3,
    mark=triangle,
    mark options={fill=mygreen,draw=mygreen,line width=1.5pt},
]
    table [
        x=X, y=Y
    ] {
        X   Y   
        0       0.43   
        1       0.50 
        2       0.53 
        3       0.50  
        4       0.50  
        5       0.43  
        6       0.47  
        7       0.50 
    };
    \addlegendentry[scale=1]{{3k Data}};
     \draw[-,dashed, thick, black, line width=1pt] 
        (axis cs:0,0.6) -- (axis cs:7,0.6);
\end{axis}

    }

 \node [anchor=center] at (12.2cm,-1.2cm) {\scalebox{1.4}{(c) AIME-25}};

\end{tikzpicture}

%% file: tables/main-results.tex










\begin{tabular}{lrcccccccccc}
\toprule[1.1pt]
\multirow{3}{*}{Method} & \multirow{3}{*}{Params.} & \multicolumn{5}{c}{Pass@128} & \multicolumn{5}{c}{Avg@128} \\ \cmidrule(l){3-7} \cmidrule(l){8-12}
& & \multicolumn{2}{c}{\textsc{AIME}} & \multicolumn{1}{c}{\textsc{BrUMO}}    & \multicolumn{2}{c}{\textsc{HMMT}}  & \multicolumn{2}{c}{\textsc{AIME}} & \multicolumn{1}{c}{\textsc{BrUMO}}    & \multicolumn{2}{c}{\textsc{HMMT}}    \\ \cmidrule(l){3-7}  \cmidrule(l){8-12}
&   & 24    & 25 & \textsc{Def.}  & \textsc{Feb.}  & \textsc{Nov.}    & 24    & 25 & \textsc{Def.}  & \textsc{Feb.}  & \textsc{Nov.}    \\  

\midrule
\rowcolor{blue!10}
\multicolumn{12}{c}{\textit{OpenR1-3k}}  \\ 
\midrule

Qwen2.5-Math & 1.5B & {46.67} & {46.67} & 43.33 & {23.33} & {40.00} & 4.30 & {3.10} & 4.09 & 0.34 & 3.15 \\
\quad +GSPO  & 1.5B & {50.00} & 36.67 & 46.67 & {23.33} & {36.67} & {6.41} & {2.86} & {11.51} & {0.42} & {3.98} \\
\quad +DAPO  & 1.5B & 43.33 & {43.33}& {56.67} & {23.33} & 33.33 & {6.64} & 2.11 & {10.70} & 0.34 & {4.04} \\ \cdashline{1-12}[4pt/1.2pt]
\quad +GRPO  & 1.5B & \textbf{43.33} & {43.33} & 53.33 & \textbf{{20.00}} & 23.33 & \textbf{4.56} & \textbf{2.76} & \textbf{5.00} & 0.23 & \textbf{3.02} \\
\quad +SPS& 1.5B & \textbf{43.33} & \textbf{{46.67}} & \textbf{{63.33}} & \textbf{{20.00}} & \textbf{{36.67}} & 4.48 & 2.61 & 4.71& \textbf{{0.36}} & 2.99  \\
\midrule
Qwen2.5-Math & 7B   & 43.33 & 43.33 & 43.33 & 13.33 & 30.00 & 3.80 & 5.50 & 1.48 & 0.16 & 1.07 \\
\quad +GSPO  & 7B   & {63.33} & 46.67 & 50.00 & 26.67 & {36.67} & {16.12} & {8.75} & 12.16 & {1.33} & 4.92 \\
\quad +DAPO  & 7B   & {63.33} & 46.67 & {56.67} & 26.67 & 26.67 & 10.39 & 5.26 & 9.74 & 0.47 & 3.65 \\ \cdashline{1-12}[4pt/1.2pt]
\quad +GRPO  & 7B   & \textbf{{70.00}} & \textbf{{50.00}} & 50.00 & \textbf{{33.33}} & {36.67} & 15.31 & \textbf{{9.48}} & \textbf{{14.84}} & \textbf{{1.48}} & {4.97} \\
\quad +SPS& 7B   & \textbf{{70.00}} & \textbf{{50.00}} & \textbf{{56.67}} & {30.00} & \textbf{{43.33}} & \textbf{{16.38} }& 8.49 & {13.85} & 1.04 & \textbf{{5.60}} \\

\midrule
\rowcolor{green!10}
\multicolumn{12}{c}{\textit{OpenR1-5k}}     \\ 
\midrule
Qwen2.5-Math & 1.5B &  46.67 & {46.67} & {43.33} & 23.33 & {40.00} & 4.30 & {3.10} & 4.09 & {0.34} & 3.15 \\
\quad +GSPO  & 1.5B & {56.67} & {46.67} & {56.67} & {26.67} & {36.67} & {6.12} & {3.41} & {11.48} & 0.42 & {4.04} \\
\quad +DAPO  & 1.5B & {53.33} & 33.33 & 50.00 & {33.33} & {40.00} & {7.45} & 1.62 & {10.86} & {0.44} & {3.67} \\ \cdashline{1-12}[4pt/1.2pt]
\quad +GRPO  & 1.5B & 43.33 & 33.33 & 50.00 & \textbf{{26.67}} & 26.67 & 4.35 & 2.69 & \textbf{4.61} & 0.29 & \textbf{3.18} \\
\quad +SPS& 1.5B & \textbf{50.00} & \textbf{{53.33}} & \textbf{{56.67}} & \textbf{{26.67}} & \textbf{{40.00}} & \textbf{4.77} & \textbf{2.94} & 4.30 & \textbf{{0.34}} & 3.10 \\
\midrule
Qwen2.5-Math & 7B   & 43.33 & 43.33 & 43.33 & 13.33 & 30.00 & 3.80 & 5.50 & 1.48 & 0.16 & 1.07 \\
\quad +GSPO  & 7B   & {66.67} & {50.00} & 50.00 & {26.67} & {40.00} & {16.25} & {9.85} & {12.68} & 1.02 & {5.34} \\
\quad +DAPO  & 7B   & {66.67} & 46.67 & 56.67 & {33.33} & 26.67 & 14.37 & 6.20 & {12.79} & {1.20} & {4.51} \\ \cdashline{1-12}[4pt/1.2pt]
\quad +GRPO  & 7B   & \textbf{{73.33}} & 46.67 & \textbf{{63.33}} & \textbf{{33.33}} & \textbf{{36.67}} & \textbf{{17.37}} & 7.66 & \textbf{11.67} & \textbf{{1.22}} & \textbf{{5.34}} \\
\quad +SPS& 7B   & 63.33 & \textbf{{53.33}} & {60.00} & \textbf{{33.33}} & 33.33 & 8.91 & \textbf{{8.41}}& 7.60 & 0.60 & 2.29 \\

\midrule
\rowcolor{yellow!40}
\multicolumn{12}{c}{\textit{OpenR1-10k}}   \\ 
\midrule
Qwen2.5-Math & 1.5B & 46.67 & {46.67} & 43.33 & {23.33} & {40.00} & 4.30 & {3.10} & 4.09 & 0.34 & 3.15 \\
\quad +GSPO  & 1.5B & {50.00} & {50.00} & {50.00} & {23.33} & 30.00 & 5.94 & 2.76 & {11.15} & {0.52} & {3.80} \\
\quad +DAPO  & 1.5B & {50.00} & 43.33 & {50.00} & {30.00} & {36.67} & {6.07} & 2.05 & {9.69} & {0.47} & {3.93} \\ \cdashline{1-12}[4pt/1.2pt]
\quad +GRPO  & 1.5B & 40.00 & 36.66 & \textbf{{56.67}} & 20.00 & \textbf{33.33} & 4.27 & \textbf{{2.84}} & \textbf{4.35} & \textbf{0.39} & 2.79 \\
\quad +SPS& 1.5B & \textbf{{56.67}} & \textbf{{50.00}} & \textbf{{56.67}} & \textbf{{30.00}} & \textbf{33.33 }& \textbf{{9.48}} & 2.53 & 3.46 & 0.29 & \textbf{3.20} \\
\midrule
Qwen2.5-Math & 7B   & 43.33 & 43.33 & 43.33 & 13.33 & 30.00 & 3.80 & 5.50 & 1.48 & 0.16 & 1.07 \\
\quad +GSPO  & 7B   & {70.00} & 46.67 & {56.67} & {26.67} & {40.00} & {15.60} & {9.48} & {13.13} & {1.33} & {4.82} \\
\quad +DAPO  & 7B   & 60.00 & 43.33 & 53.33 & {26.67} & 30.00 & {14.82} & 4.85 & 12.93 & {1.09} & 4.35 \\ \cdashline{1-12}[4pt/1.2pt]
\quad +GRPO  & 7B   & \textbf{{66.67}} & \textbf{{56.67}} & 50.00 & {30.00} & {33.33} & \textbf{14.69}& \textbf{{8.75}} & \textbf{{13.41}} & \textbf{1.02} & \textbf{{5.21}} \\
\quad +SPS& 7B & 63.33 & {53.33} & \textbf{{66.67}} & \textbf{{36.67}} & \textbf{{36.67}} & 13.10 & 8.17 & 10.05 & 0.86 & 4.56 \\

\bottomrule[1.1pt]
\end{tabular}

%% file: images/ablation.tex
\definecolor{blue1}{RGB}{29,143,255}
\definecolor{blue2}{RGB}{74,165,255}
\definecolor{blue3}{RGB}{142,199,255}
\definecolor{mygreen}{RGB}{46,139,87}
\definecolor{myred}{RGB}{255,152,150}
\definecolor{myblue}{RGB}{30,144,255}

\begin{tikzpicture}

    \scriptsize{    
    \begin{axis}
        [
        ymajorgrids,
        xmajorgrids,
        grid style=dashed,
        at={(10.4cm,0)},
        height=.30\textwidth,
        width=.47\textwidth,
        xmin=0.9,
        xmax=5.1,
        xtick={1,2,...,5},
        xticklabels={1,2,...,5}, 
        xticklabel style={/pgf/number format/fixed}, 
        scaled x ticks=false,
        ymin=-0.05,
        ymax=0.75,
        ytick={0,0.1,0.2,...,0.6,0.7},
        xtick=data,
        x tick label style={/pgf/number format/fixed,
            /pgf/number format/fixed zerofill,
        /pgf/number format/precision=1, scale=1.2},
        y tick label style={/pgf/number format/fixed, xshift=-0.5ex,
            /pgf/number format/fixed zerofill,
        /pgf/number format/precision=1, scale=1.2},
        ylabel=\scriptsize{Pass@128},
        ylabel style={yshift=-3.0ex,scale=1.2},
        xlabel=\scriptsize{Sampling Size},
        xlabel style={yshift=1ex,scale=1.2},
        legend style = {
                        fill=white,        
                        fill opacity=0.4,  
                        draw opacity=0.5,  
                        text opacity=1,
                        column sep=0.05em,
                        row sep=-0.1em,
                        draw=black!30,
                        inner sep=0.5pt,
                        yshift=0.5ex,
                        legend columns=1},
        legend pos=north west
                        ]

    \addplot+[
    myblue,
    mark=none,
    mark size=1.2pt,
    line width=1.4pt,
    smooth, tension=0.3,
    mark=*,
    mark options={fill=myblue,draw=myblue,line width=1.5pt},
]
    table [
        x=X, y=Y
    ] {
        X   Y   
        1       0.5333 
        2       0.4667
        3       0.5 
        4       0.5  
        5       0.5  
    };
    \addlegendentry[scale=0.8]{{\textsc{BrUMO}-25}};

    \addplot+[
    myred,
    mark=none,
    mark size=1.2pt,
    line width=1.4pt,
    smooth, tension=0.3,
    mark=square,
    mark options={fill=myred,draw=myred,line width=1.5pt},
]
    table [
        x=X, y=Y
    ] {
        X   Y   
        1       0.2333 
        2       0.1333 
        3       0.2333  
        4       0.2333 
        5       0.1667 
    };
    \addlegendentry[scale=0.8]{\textsc{HMMT-Feb-25}};

     \addplot+[
    mygreen,
    mark=none,
    mark size=1.2pt,
    line width=1.4pt,
    smooth, tension=0.3,
    mark=triangle,
    mark options={fill=mygreen,draw=mygreen,line width=1.5pt},
]
    table [
        x=X, y=Y
    ] {
        X   Y   
        1       0.2667 
        2       0.2667 
        3       0.3  
        4       0.3  
        5       0.3  
    };
    \addlegendentry[scale=0.8]{\textsc{HMMT-Nov-25}};
\end{axis}
    }

\end{tikzpicture}

%% file: algorithm/looped.tex
\begin{algorithm*}[h]
\caption{Iterative SPS}
\label{alg:looped}
\KwIn{Base policy $\pi_{\theta_0}(\cdot)$, dataset $D$, group size $n$, sampling size $k$}
\KwOut{Enhanced policy $\pi_{\theta}(\cdot)$}

Initialize policy $\pi_{\theta}(\cdot) \leftarrow \pi_{\theta_0}(\cdot)$\;

\While{not converged}{
    \tcp{Stage 1: Vanilla RL}
    Update $\pi_{\theta}(\cdot)$ on $D$ using vanilla RL to encourage exploration\;
    Collect grouped rollouts:
    \[
    Y = \{\, y_x^1, \dots, y_x^n \mid y_x^i \sim \pi_\theta(\cdot \mid x), x \in D \,\}
    \]
    
    \tcp{PL2TE}
    Sample a subset $Y' \subset Y$, emphasizing low-likelihood or under-explored trajectories\;
    \vspace{2mm}
    \tcp{Stage 2: IRL}
    Update $\pi_{\theta}(\cdot)$ via IRL on $Y'$ by minimizing $\mathcal{L}_{\mathrm{IRL}}$\;
    \vspace{2mm}
    \tcp{Reshape the policy distribution to mitigate probability squeezing}
}

\Return{$\pi_{\theta}$}\;
\end{algorithm*}

%% file: tables/hyperparameters.tex

\begin{tabular}{lll}
\toprule[1.1pt]
Method & Parameter Name & Value \\ \cmidrule(r){1-3}
GRPO & beta & 0.01 \\
 & group\_size & 8 \\ \cmidrule(r){1-3}
DAPO & epsilon\_high & 0.28 \\
 & max\_resample\_times & 3 \\ 
  & soft\_cache\_length & 2048 \\ \cmidrule(r){1-3}
GSPO & beta & 0.0 \\
 & epsilon & 3e-4 \\
& epsilon\_high & 4e-4 \\
 & steps\_per\_generation & 4 \\

\bottomrule[1.1pt]
\end{tabular}

%% file: images/appendix_acc_bucket.tex
\definecolor{blue1}{RGB}{29,143,255}
\definecolor{blue2}{RGB}{74,165,255}
\definecolor{blue3}{RGB}{142,199,255}
\definecolor{mygreen}{RGB}{46,139,87}
\definecolor{myred}{RGB}{255,152,150}
\definecolor{myblue}{RGB}{30,144,255}

\begin{tikzpicture}
       \scriptsize{    
\begin{axis}[
    ybar, 
    bar width=7pt,
    ymajorgrids,
    xmajorgrids,
    grid style=dashed,
    at={(0,0)},
    height=.27\textwidth,
    width=.28\textwidth,
    xmin=-0.1,
    xmax=1.1,
    ymin=0,
    ymax=100,
    xtick={0, 0.1, 0.2, 0.3, 0.4, 0.5, 0.6, 0.7, 0.8, 0.9, 1.0},
    xticklabels={0, , 0.2, , 0.4, , 0.6, , 0.8, , 1.0, }, 
    xticklabel style={/pgf/number format/fixed, font=\tiny, scale=1.2},
    xtick pos=bottom,
    major tick length=2pt,
    ytick={0, 20, 40, 60, 80, 100},
    scaled x ticks=false,
    ylabel=\scriptsize{Count},
    ylabel style={yshift=-6.0ex,scale=1.2},
    xlabel=\scriptsize{Accuracy Interval},
    xlabel style={yshift=1ex,scale=1.2},
    legend style={column sep=.45em, draw=black!30, legend columns=1},
    legend pos=north east
    ]
    
    \addplot+[ybar, fill=blue!15, draw=blue!40!black] coordinates {
        (0.0, 90)
        (0.1, 41)
        (0.2, 14)
        (0.3, 2)
        (0.4, 3)
        (0.5, 0)
        (0.6, 0)
        (0.7, 0)
        (0.8, 0)
        (0.9, 0)
        (1.0, 0)
        (1.1, 0)
    };
    \addplot+[
        sharp plot,       
        draw=blue!60,      
        fill=none,
        line width=1.0pt, 
        mark=*,           
        mark options={fill=blue!80, draw=none, scale=0.4} 
    ] coordinates {
        (0.0, 90)
        (0.1, 41)
        (0.2, 14)
        (0.3, 2)
        (0.4, 3)
        (0.5, 0)
        (0.6, 0)
        (0.7, 0)
        (0.8, 0)
        (0.9, 0)
        (1.0, 0)
        (1.1, 0)
    };
\end{axis}

\begin{axis}[
    ybar,
    bar width=7pt,
    ymajorgrids,
    xmajorgrids,
    grid style=dashed,
    at={(3.9cm,0)}, 
    height=.27\textwidth,
    width=.28\textwidth,
    xmin=-0.1,
    xmax=1.1,
    ymin=0,
    ymax=100,
    xtick={0, 0.1, 0.2, 0.3, 0.4, 0.5, 0.6, 0.7, 0.8, 0.9, 1.0},
    xticklabels={0, , 0.2, , 0.4, , 0.6, , 0.8, , 1.0, },
    xticklabel style={/pgf/number format/fixed, font=\tiny, scale=1.2},
    xtick pos=bottom,       
    major tick length=2pt,  
    ytick={0, 20, 40, 60, 80, 100},
    scaled x ticks=false,
    ylabel=\scriptsize{Count},
    ylabel style={yshift=-6.0ex,scale=1.2},
    xlabel=\scriptsize{Accuracy Interval},
    xlabel style={yshift=1ex,scale=1.2},
    legend style={column sep=.45em, draw=black!30, legend columns=1},
    legend pos=north east
    ]
    
    \addplot+[ybar, fill=blue!15, draw=blue!40!black] coordinates {
        (0.0, 90)
        (0.1, 40)
        (0.2, 8)
        (0.3, 6)
        (0.4, 1)
        (0.5, 1)
        (0.6, 3)
        (0.7, 0)
        (0.8, 1)
        (0.9, 0)
        (1.0, 0)
        (1.1, 0)
    };
    \addplot+[
    sharp plot,       
    draw=blue!60,      
    fill=none,
    line width=1.0pt, 
    mark=*,           
    mark options={fill=blue!80, draw=none, scale=0.4} 
] coordinates {
        (0.0, 90)
        (0.1, 40)
        (0.2, 8)
        (0.3, 6)
        (0.4, 1)
        (0.5, 1)
        (0.6, 3)
        (0.7, 0)
        (0.8, 1)
        (0.9, 0)
        (1.0, 0)
        (1.1, 0)
};

\end{axis}
}

    \scriptsize{    
\begin{axis}[
    ybar, 
    bar width=7pt,
    ymajorgrids,
    xmajorgrids,
    grid style=dashed,
    at={(7.8cm,0)},
    height=.27\textwidth,
    width=.28\textwidth,
    xmin=-0.1,
    xmax=1.1,
    ymin=0,
    ymax=100,
    xtick={0, 0.1, 0.2, 0.3, 0.4, 0.5, 0.6, 0.7, 0.8, 0.9, 1.0},
    xticklabels={0, , 0.2, , 0.4, , 0.6, , 0.8, , 1.0, }, 
    xticklabel style={/pgf/number format/fixed, font=\tiny, scale=1.2},
    xtick pos=bottom,
    major tick length=2pt,
    ytick={0, 20, 40, 60, 80, 100},
    scaled x ticks=false,
    ylabel=\scriptsize{Count},
    ylabel style={yshift=-6.0ex,scale=1.2},
    xlabel=\scriptsize{Accuracy Interval},
    xlabel style={yshift=1ex,scale=1.2},
    legend style={column sep=.45em, draw=black!30, legend columns=1},
    legend pos=north east
    ]
    
    \addplot+[ybar, fill=blue!15, draw=blue!40!black] coordinates {
        (0.0, 98)
        (0.1, 44)
        (0.2, 3)
        (0.3, 1)
        (0.4, 2)
        (0.5, 1)
        (0.6, 0)
        (0.7, 0)
        (0.8, 1)
        (0.9, 0)
        (1.0, 0)
        (1.1, 0)
    };
    \addplot+[
        sharp plot,       
        draw=blue!60,      
        fill=none,
        line width=1.0pt, 
        mark=*,           
        mark options={fill=blue!80, draw=none, scale=0.4} 
    ] coordinates {
        (0.0, 98)
        (0.1, 44)
        (0.2, 3)
        (0.3, 1)
        (0.4, 2)
        (0.5, 1)
        (0.6, 0)
        (0.7, 0)
        (0.8, 1)
        (0.9, 0)
        (1.0, 0)
    };
\end{axis}
\begin{axis}[
    ybar, 
    bar width=7pt,
    ymajorgrids,
    xmajorgrids,
    grid style=dashed,
    at={(11.7cm,0)},
    height=.27\textwidth,
    width=.28\textwidth,
    xmin=-0.1,
    xmax=1.1,
    ymin=0,
    ymax=100,
    xtick={0, 0.1, 0.2, 0.3, 0.4, 0.5, 0.6, 0.7, 0.8, 0.9, 1.0},
    xticklabels={0, , 0.2, , 0.4, , 0.6, , 0.8, , 1.0, }, 
    xticklabel style={/pgf/number format/fixed, font=\tiny, scale=1.2},
    xtick pos=bottom,
    major tick length=2pt,
    ytick={0, 20, 40, 60, 80, 100},
    scaled x ticks=false,
    ylabel=\scriptsize{Count},
    ylabel style={yshift=-6.0ex,scale=1.2},
    xlabel=\scriptsize{Accuracy Interval},
    xlabel style={yshift=1ex,scale=1.2},
    legend style={column sep=.45em, draw=black!30, legend columns=1},
    legend pos=north east
    ]
    
    \addplot+[ybar, fill=blue!15, draw=blue!40!black] coordinates {
        (0.0, 98)
        (0.1, 44)
        (0.2, 3)
        (0.3, 1)
        (0.4, 2)
        (0.5, 1)
        (0.6, 0)
        (0.7, 0)
        (0.8, 1)
        (0.9, 0)
        (1.0, 0)
        (1.1, 0)
    };
    \addplot+[
        sharp plot,       
        draw=blue!60,      
        fill=none,
        line width=1.0pt, 
        mark=*,           
        mark options={fill=blue!80, draw=none, scale=0.4} 
    ] coordinates {
        (0.0, 98)
        (0.1, 44)
        (0.2, 3)
        (0.3, 1)
        (0.4, 2)
        (0.5, 1)
        (0.6, 0)
        (0.7, 0)
        (0.8, 1)
        (0.9, 0)
        (1.0, 0)
    };
\end{axis}
    \node [anchor=center] at (7.0cm,-1.6cm) {\scalebox{1.4}{(a)Accuracy distribution of Qwen2.5-Math-1.5B. From left to right: base model, GRPO, DAPO, and GSPO.}};
    }

         \scriptsize{    
\begin{axis}[
    ybar, 
    bar width=7pt,
    ymajorgrids,
    xmajorgrids,
    grid style=dashed,
    at={(0,-5cm)},
    height=.27\textwidth,
    width=.28\textwidth,
    xmin=-0.1,
    xmax=1.1,
    ymin=0,
    ymax=100,
    xtick={0, 0.1, 0.2, 0.3, 0.4, 0.5, 0.6, 0.7, 0.8, 0.9, 1.0},
    xticklabels={0, , 0.2, , 0.4, , 0.6, , 0.8, , 1.0, }, 
    xticklabel style={/pgf/number format/fixed, font=\tiny, scale=1.2},
    xtick pos=bottom,
    major tick length=2pt,
    ytick={0, 20, 40, 60, 80, 100},
    scaled x ticks=false,
    ylabel=\scriptsize{Count},
    ylabel style={yshift=-6.0ex,scale=1.2},
    xlabel=\scriptsize{Accuracy Interval},
    xlabel style={yshift=1ex,scale=1.2},
    legend style={column sep=.45em, draw=black!30, legend columns=1},
    legend pos=north east
    ]
    
    \addplot+[ybar, fill=blue!15, draw=blue!40!black] coordinates {
        (0.0, 90)
        (0.1, 41)
        (0.2, 14)
        (0.3, 2)
        (0.4, 3)
        (0.5, 0)
        (0.6, 0)
        (0.7, 0)
        (0.8, 0)
        (0.9, 0)
        (1.0, 0)
        (1.1, 0)
    };
    \addplot+[
        sharp plot,       
        draw=blue!60,      
        fill=none,
        line width=1.0pt, 
        mark=*,           
        mark options={fill=blue!80, draw=none, scale=0.4} 
    ] coordinates {
        (0.0, 90)
        (0.1, 41)
        (0.2, 14)
        (0.3, 2)
        (0.4, 3)
        (0.5, 0)
        (0.6, 0)
        (0.7, 0)
        (0.8, 0)
        (0.9, 0)
        (1.0, 0)
        (1.1, 0)
    };
\end{axis}

\begin{axis}[
    ybar,
    bar width=7pt,
    ymajorgrids,
    xmajorgrids,
    grid style=dashed,
    at={(3.9cm,-5cm)}, 
    height=.27\textwidth,
    width=.28\textwidth,
    xmin=-0.1,
    xmax=1.1,
    ymin=0,
    ymax=100,
    xtick={0, 0.1, 0.2, 0.3, 0.4, 0.5, 0.6, 0.7, 0.8, 0.9, 1.0},
    xticklabels={0, , 0.2, , 0.4, , 0.6, , 0.8, , 1.0, },
    xticklabel style={/pgf/number format/fixed, font=\tiny, scale=1.2},
    xtick pos=bottom,       
    major tick length=2pt,  
    ytick={0, 20, 40, 60, 80, 100},
    scaled x ticks=false,
    ylabel=\scriptsize{Count},
    ylabel style={yshift=-6.0ex,scale=1.2},
    xlabel=\scriptsize{Accuracy Interval},
    xlabel style={yshift=1ex,scale=1.2},
    legend style={column sep=.45em, draw=black!30, legend columns=1},
    legend pos=north east
    ]
    
    \addplot+[ybar, fill=blue!15, draw=blue!40!black] coordinates {
        (0.0, 90)
        (0.1, 40)
        (0.2, 8)
        (0.3, 6)
        (0.4, 1)
        (0.5, 1)
        (0.6, 3)
        (0.7, 0)
        (0.8, 1)
        (0.9, 0)
        (1.0, 0)
        (1.1, 0)
    };
    \addplot+[
    sharp plot,       
    draw=blue!60,      
    fill=none,
    line width=1.0pt, 
    mark=*,           
    mark options={fill=blue!80, draw=none, scale=0.4} 
] coordinates {
        (0.0, 90)
        (0.1, 40)
        (0.2, 8)
        (0.3, 6)
        (0.4, 1)
        (0.5, 1)
        (0.6, 3)
        (0.7, 0)
        (0.8, 1)
        (0.9, 0)
        (1.0, 0)
        (1.1, 0)
};

\end{axis}
}

    \scriptsize{    
\begin{axis}[
    ybar, 
    bar width=7pt,
    ymajorgrids,
    xmajorgrids,
    grid style=dashed,
    at={(7.8cm,-5cm)},
    height=.27\textwidth,
    width=.28\textwidth,
    xmin=-0.1,
    xmax=1.1,
    ymin=0,
    ymax=100,
    xtick={0, 0.1, 0.2, 0.3, 0.4, 0.5, 0.6, 0.7, 0.8, 0.9, 1.0},
    xticklabels={0, , 0.2, , 0.4, , 0.6, , 0.8, , 1.0, }, 
    xticklabel style={/pgf/number format/fixed, font=\tiny, scale=1.2},
    xtick pos=bottom,
    major tick length=2pt,
    ytick={0, 20, 40, 60, 80, 100},
    scaled x ticks=false,
    ylabel=\scriptsize{Count},
    ylabel style={yshift=-6.0ex,scale=1.2},
    xlabel=\scriptsize{Accuracy Interval},
    xlabel style={yshift=1ex,scale=1.2},
    legend style={column sep=.45em, draw=black!30, legend columns=1},
    legend pos=north east
    ]
    
    \addplot+[ybar, fill=blue!15, draw=blue!40!black] coordinates {
        (0.0, 92)
        (0.1, 38)
        (0.2, 5)
        (0.3, 7)
        (0.4, 3)
        (0.5, 0)
        (0.6, 4)
        (0.7, 0)
        (0.8, 0)
        (0.9, 1)
        (1.0, 0)
        (1.1, 0)
    };
    \addplot+[
        sharp plot,       
        draw=blue!60,      
        fill=none,
        line width=1.0pt, 
        mark=*,           
        mark options={fill=blue!80, draw=none, scale=0.4} 
    ] coordinates {
        (0.0, 92)
        (0.1, 38)
        (0.2, 5)
        (0.3, 7)
        (0.4, 3)
        (0.5, 0)
        (0.6, 4)
        (0.7, 0)
        (0.8, 0)
        (0.9, 1)
        (1.0, 0)
        (1.1, 0)
    };
\end{axis}
\begin{axis}[
    ybar, 
    bar width=7pt,
    ymajorgrids,
    xmajorgrids,
    grid style=dashed,
    at={(11.7cm,-5cm)},
    height=.27\textwidth,
    width=.28\textwidth,
    xmin=-0.1,
    xmax=1.1,
    ymin=0,
    ymax=100,
    xtick={0, 0.1, 0.2, 0.3, 0.4, 0.5, 0.6, 0.7, 0.8, 0.9, 1.0},
    xticklabels={0, , 0.2, , 0.4, , 0.6, , 0.8, , 1.0, }, 
    xticklabel style={/pgf/number format/fixed, font=\tiny, scale=1.2},
    xtick pos=bottom,
    major tick length=2pt,
    ytick={0, 20, 40, 60, 80, 100},
    scaled x ticks=false,
    ylabel=\scriptsize{Count},
    ylabel style={yshift=-6.0ex,scale=1.2},
    xlabel=\scriptsize{Accuracy Interval},
    xlabel style={yshift=1ex,scale=1.2},
    legend style={column sep=.45em, draw=black!30, legend columns=1},
    legend pos=north east
    ]
    
    \addplot+[ybar, fill=blue!15, draw=blue!40!black] coordinates {
        (0.0, 78)
        (0.1, 40)
        (0.2, 8)
        (0.3, 4)
        (0.4, 8)
        (0.5, 3)
        (0.6, 1)
        (0.7, 5)
        (0.8, 1)
        (0.9, 2)
        (1.0, 0)
        (1.1, 0)
    };
    \addplot+[
        sharp plot,       
        draw=blue!60,      
        fill=none,
        line width=1.0pt, 
        mark=*,           
        mark options={fill=blue!80, draw=none, scale=0.4} 
    ] coordinates {
         (0.0, 78)
        (0.1, 40)
        (0.2, 8)
        (0.3, 4)
        (0.4, 8)
        (0.5, 3)
        (0.6, 1)
        (0.7, 5)
        (0.8, 1)
        (0.9, 2)
        (1.0, 0)
        (1.1, 0)
    };
\end{axis}
    \node [anchor=center] at (7.0cm,-6.6cm) {\scalebox{1.4}{(b)Accuracy distribution of Qwen2.5-Math-7B. From left to right: base model, GRPO, DAPO, and GSPO.}};
    }

\end{tikzpicture}

%% file: tables/additional-res.tex
\begin{tabular}{lllllllll}
\toprule[1.1pt]
Method                        & \textsc{AIME-24} & \textsc{AIME-25} & \textsc{BrUMO} & \textsc{HMMT-Feb} & \textsc{HMMT-Nov} & \textsc{MATH-500} & \textsc{MMLU}  & \textsc{GPQA-Diamond} \\ 
\midrule
DeepSeek-R1-Distill-Qwen-1.5B & 63.33   & 36.67   & 60.00 & 33.33    & 36.67    & 72.39    & 44.33 & 14.14        \\
+GRPO                         & 70.00   & \underline{46.67}   & \textbf{80.00} & \underline{40.00}    & 36.67    & 83.42    & 44.86 & 19.70        \\
+GSPO                         & \underline{73.33}   & \textbf{50.00}   & \underline{66.67} & 30.00    & \underline{40.00}    & 76.51    & 44.16 & 21.21        \\
+DAPO                         & 66.67   & \underline{40.00}   & 60.00 & 33.33    & 26.67    & \underline{83.46}    & \textbf{46.26} & \textbf{33.33}        \\
+SPS                          & \textbf{80.00}   & \underline{46.67}   & \textbf{80.00} & \textbf{46.67}    & \textbf{43.33}    & \textbf{84.25}    & \underline{46.13} & \underline{29.80}  \\
\bottomrule[1.1pt]
\end{tabular}

%% file: tables/div-analysis.tex
\begin{tabular}{llll}
\toprule[1.1pt]
Stage     & BASE   & GRPO  & SPS \\ \midrule
Similarity & 88.34  & 88.18 & 86.82\\
\bottomrule[1.1pt]
\end{tabular}

%% file: tables/cost-analysis.tex
\begin{tabular}{lll}
\toprule[1.1pt]
Stage     & RL    & IRL  \\ \midrule
Time(min) & 68.10 & 2.25 \\
\bottomrule[1.1pt]
\end{tabular}

%% file: tables/squeeze-analysis.tex
\begin{tabular}{lll}
\toprule[1.1pt]
Method     & AIME-24    & AIME-25  \\ \midrule
Base & -124.2745537 & -120.7128742 \\
GRPO & -122.4005783 & -115.0455817 \\
GSPO & -121.8923759 & -115.0455817 \\
SPS  & -124.3440186 & -115.0455817 \\
\bottomrule[1.1pt]
\end{tabular}